\pdfoutput=1

\documentclass[11pt]{article}

\usepackage{acl}
\usepackage{times}
\usepackage{latexsym}

\usepackage[T1]{fontenc}
\usepackage{wrapfig}
\usepackage{booktabs}       
\usepackage{amsfonts}       
\usepackage{nicefrac}       
\usepackage{microtype}      
\usepackage{xcolor}         
\usepackage{graphicx}

\usepackage[utf8]{inputenc}

\usepackage{microtype}

\usepackage{inconsolata}
\usepackage{todonotes} 

\title{AUTODIAL: Efficient Asynchronous Task-Oriented Dialogue Model}

\author{
Prajjwal Bhargava \hspace{6mm} Pooyan Amini  \hspace{6mm} Shahin Shayandeh \hspace{6mm}  Chinnadhurai Sankar \\
Meta AI \\
\small{\texttt{prajj@meta.com}}
}

\begin{document}
\maketitle
\begin{abstract}
As large dialogue models become commonplace in practice, the problems surrounding high compute requirements for training, inference, and larger memory footprint persist. In this work, we present \textsc{autodial}, a multi-task dialogue model that aims to address these challenges concerned with the deployment of such models. \textsc{Autodial} extends the BART architecture with parallel decoders to perform the four key dialogue tasks. Using classification decoders over generative ones allows \textsc{autodial} to significantly reduce memory footprint (11x fewer parameters) and achieve faster inference (3-6x speedups) times compared to an existing competitive generative approach which relies on the text-to-text interface such as \textsc{simpletod}. Our results show that extending current dialogue models with parallel decoders can be a viable alternative for deployment in resource-constrained environments.
\end{abstract}

\section{Introduction}
Dialogue systems are playing an increasingly growing role in how humans interact with devices and services. Existing works such as SimpleTOD \cite{NEURIPS2020_e9462095} have adopted generative (text-to-text) approaches which require conditioning upon different text sequences for key dialogue tasks such as action/response generation and dialogue state tracking (DST). While these approaches have managed to achieve competitive results on numerous tasks, they often have large memory requirements and prohibitively high inference latency, making them impractical for use in resource-restricted environments. Moreover, accounting for the ever-growing changing needs of the user such as adding support for a new dialogue task or a domain in a modular fashion, is non-trivial with a generative approach.

\begin{figure}
    \centering
      \includegraphics{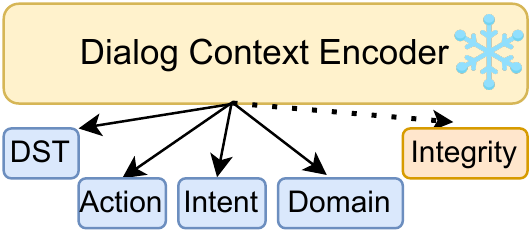}
    \caption{\textsc{Autodial} architecture. Representations formed from the encoder are leveraged by parallel decoders (shown in blue and orange) for several dialogue tasks, indicated by arrows. Dialog Context Encoder is kept frozen. All the tasks highlighted in blue were considered as part of this study. More tasks can be added by solely training the task-specific decoder while keeping everything else fixed i.e non-trainable (highlighted in orange).}
    \label{fig:autodial}
    \vspace{-2em}
\end{figure}

To address these issues, we present \textsc{Autodial}, an \textbf{a}synchrono\textbf{u}s \textbf{t}ask-\textbf{o}riented \textbf{dial}ogue model that can perform inference on multiple dialogue tasks concurrently in an asynchronous manner. The model, as shown in \autoref{fig:autodial}, extends the popular BART architecture \cite{lewis-etal-2020-bart} by adding parallel decoders for dialogue tasks allowing it to achieve 3-6x faster inference times while having 11x fewer parameters with no degradation in performance compared to \textsc{simpletod} on three key dialogue tasks of \textsc{multiwoz} and \textsc{google sgd} each. Additionally, this architecture enables the flexible addition of new decoders in an asynchronous manner i.e training an additional small decoder without training any other component of the model, making it easier to support new dialogue tasks or domains\footnote{https://colinraffel.com/blog/a-call-to-build-models-like-we-build-open-source-software.html}. Since the pre-trained representations play a cardinal role, we also explore pre-finetuning of the \textsc{Autodial} and find that pre-finetuning benefits the larger variant marginally over smaller variant suggesting that it becomes more beneficial as the model size increases.
We open source our code here\footnote{https://github.com/prajjwal1/autodial}.

\section{Background}
In this section, we focus our discussion on \textsc{simpletod}, which formed the primary baseline for comparison with \textsc{autodial} along with pre-finetuning  which was done to study its role on \textsc{autodial}. Additionally we provide details on tasks which were considered for training and evaluating \textsc{autodial}.

\paragraph{SimpleTOD:} Conventionally task-oriented dialogue (TOD) systems have been built independently with separate supervision of each component. A popular approach \textsc{simpletod} which uses an identical architecture as BART \cite{lewis-etal-2020-bart} recasts the TOD as a causal unidirectional language modeling task to solve numerous sub-tasks in a text-to-text manner. By training the model end-to-end to generate a specific type of response when provided with a specific type of context, the user can generate belief states, actions and de-lexicalized response as part of a multi-step process. The multi-step process includes generating a belief state from system and user utterances, in the next step, model's output from previous step is concatenated with utterances to generate actions and so on.

\textbf{Pre-finetuning} (PFT) is an additional training phase proposed by \citet{aghajanyan-etal-2021-muppet} wherein a language model undergoes multi-task training on many downstream tasks after pre-training and before fine-tuning to encode task-specific representations. We specified the datasets used for performing pre-finetuning in \autoref{sec:experimental_details}.

\paragraph{Tasks:} We consider two datasets for our study that are large-scale, multi-turn, multi-domain task-oriented dialogue datasets, namely \textsc{multiwoz v2.2} and \textsc{google sgd}. \textsc{multiwoz v2.2} contains 8.4k dialogues across seven domains, with an average of 13.46 turns per dialogue. \textsc{Google sgd} is a larger dataset with over 16k dialogues covering 21 domains, with an average of 20.4 turns per dialogue.

We briefly define the four dialogue tasks that are common across both datasets used in this study. Note that all of these tasks require the same input i.e. context $C_{t}$.

\begin{itemize}
    \item \textbf{Dialogue State Tracking} involves generating the belief state $B_{t}$, which is a list of triplets containing values for slots in a particular domain (\textsc{domain, slot\_name, value}).
    \item \textbf{Dialogue Act Prediction} requires prediction of a triplet constructed by concatenating the domain, entity, and slot key for that dialogue turn. An example dialogue act would be \texttt{"Attraction\_Inform\_address"} where attraction, inform and address are the domain, entity, and slot keys respectively. There are a total of 93 and 444 classes in \textsc{multiwoz} and \textsc{google sgd}, respectively.
    \item \textbf{Dialogue Intent Prediction} necessitates deriving the intent of the user. Examples of intents include ${\texttt{Taxi-Request, Hotel-Inform}}$. There are 18 and 23 intents in \textsc{multiwoz} and \textsc{google sgd}, respectively.
    \item \textbf{Domain Prediction} focuses on the classification of a given example into a set of the domain such as \textsc{restaurant}.

\end{itemize}

\section{Experimental Details}
\label{sec:experimental_details}

\begin{table*}
\footnotesize
\centering
  \begin{tabular}{lll}
    \toprule
     Baseline                   &  Small / Large &  Small / Large  \\
                                &     \textsc{multiwoz}       & \textsc{google sgd} \\
     \midrule
    \multicolumn{3}{c}{Dialogue State Tracking}   (JGA)                \\
    \midrule
    \textsc{simpletod (baseline)}               &   50.6$_{[0.33]}$ / 56.4$_{[0.23]}$ & 29.97$_{[0.24]}$ / 31.81$_{[0.38]}$      \\
    \textsc{Generative}             &   37.2$_{[0.4]}$ / 45.4$_{[0.26]}$ & \textbf{26.8}$_{[0.09]}$   / 29.6$_{[0.3]}$   \\

    \midrule
    \multicolumn{3}{c}{Dialogue Act Prediction} (Accuracy)    \\
    \midrule
     \textsc{simpletod (baseline)}                     & 82$_{[0.18]}$ / 84.16$_{[0.25]}$ &   51$_{[0.32]}$ / 50.6$_{[0.42]}$     \\
    \textsc{generative}                   & 76.2$_{[0.49]}$ / 71.1$_{[7.5]}$ & 47.9$_{[0.1]}$ / 45.3$_{[0.38]}$     \\
    \textsc{autodial}               & 69.6$_{[0.69]}$ / 68.5$_{[0.19]}$ & 45.8$_{[0.23]}$ / \textbf{51.8}$_{[0.53]}$     \\

    \midrule
    \multicolumn{3}{c}{Dialogue Intent Prediction}  (Accuracy)       \\
    \midrule
    \textsc{simpletod (baseline)}                 &  85$_{[0.18]}$ / 86.94$_{[0.39]}$  & 69.03$_{[0.16]}$ / 66$_{[0.34]}$    \\
    \textsc{generative}               &  82.5$_{[0.36]}$ / 82.6$_{[0.46]}$  &  68.5$_{[0.09]}$ / 66$_{[0.34]}$   \\
    \textsc{autodial}           &  \textbf{83.5}$_{[0.6]}$ / \textbf{83.6}$_{[0.24]}$  & \textbf{69.2}$_{[0.19]}$ / \textbf{67.19}$_{[0.33]}$    \\


    \midrule
    \multicolumn{3}{c}{Dialogue Domain Prediction}   (Accuracy)            \\
    \midrule
    \textsc{simpletod (baseline)}                     & 96.83$_{[0.45]}$ / 96.23$_{[0.77]}$   &  74.73$_{[0.47]}$ /  75.5$_{[0.47]}$       \\
    \textsc{generative}                  & 94.9$_{[2]}$ / 97.2$_{[0.69]}$  &  74.04$_{[0.21]}$ /  71$_{[0.86]}$      \\
    \textsc{autodial}                & \textbf{96.2}$_{[0.28]}$ / 95.6$_{[0.43]}$    &  \textbf{76}$_{[1.42]}$ / \textbf{73.5}$_{[0.23]}$         \\

    \bottomrule
  \end{tabular}

   \caption{Results on \textsc{multiwoz 2.2} and \textsc{google sgd}. Numbers in subscript denote standard deviation observed across 3 runs. Bolded numbers represent cases where \textsc{autodial} is competitive to the other two baselines namely \textsc{simpletod} and \textsc{generative}. \textsc{autodial} is either competitive with or outperforms the baselines.}
    \label{tab:quant_results}
  \end{table*}

\paragraph{Architecture:} In addition to BART architecture, \textsc{autodial} contains additional classification decoders in parallel that can work in an asynchronous fashion and are specialized for each dialogue task. The encoder representations are kept frozen for retention of pre-trained knowledge. For each of the task, the encoder receives the same context $C_{t}$, which is the concatenation of user ($U$) and system ($S$) utterances ($C_{t} = [U_{0}, S_{0}...U_{t}]$). Each parallel decoder is trained independently of each other and gets the same vector representation from the encoder. In \textsc{autodial}, the generative decoder used for the DST task is a BART decoder, while separate identical classification decoders were used for the remaining tasks which are comprised of a 2-layer transformer network. Each decoder receives its corresponding label for computation of binary cross entropy loss during training time. The independent nature of the decoder allows for outputs to be gathered in an asynchronous manner unlike generative decoder which requires multiple forward passes to produce all the outputs for the dialogue tasks.

\paragraph{Baseline}: \textsc{simpletod} served as the primary baseline and represents the conventional case where all the weights of the model are trained. In addition, we compare \textsc{autodial} with a modified variant of \textsc{simpletod} wherein the encoder weights were kept fixed but the generative decoder was fine-tuned on the respective downstream tasks, we call this baseline \textsc{generative}. This baseline allowed us to evaluate how effective the parallel classification decoders of \textsc{autodial} are against a generative decoder \cite{2020t5} as the encoder is kept common across both.

\paragraph{Metrics: } We measured performance on DST using Joint Goal Accuracy (JGA). It measures the accuracy of generated belief states against oracle belief states. Results on the classification tasks report accuracy (exact matching). We treat all classification tasks as a multi-label classification problem. It is worth noting that we performed DST in the same way as \textsc{simpletod} and has been included in this study to demonstrate that \textsc{autodial} can have generative decoders also. \textsc{generative} baseline in \autoref{tab:quant_results} is equivalent to having a generative decoder (required by DST task) in \textsc{autodial}. This study is concerned with classification tasks as a result of which the performance and memory gains described in this work apply to all tasks except DST.

\begin{table}
\footnotesize
\centering
  \begin{tabular}{lll}
    \toprule
    Model                   &  Parameter count \\
     \midrule
    \textsc{Encoder}                        &       202M \\
    \textsc{Autodial decoder} (x1)          &       26M  \\
    \textsc{simpletod decoder}               &      202.6M \\
    \textsc{simpletod}                      &       406M \\
    \bottomrule
  \end{tabular}

   \caption{Parameter count of different components used in \textsc{autodial} and \textsc{simpletod}. The encoder weights are common to \textsc{autodial} and \textsc{simpletod} and are non-trainable during fine-tuning stage for \textsc{autodial}.}
    \label{tab:param_count_decoder}
    \vspace{-2em}

  \end{table}

\begin{table*}
  \centering
\footnotesize
  \begin{tabular}{lll}

    \toprule
     Baseline                   &  Small / Large & Small / Large  \\
                                & \textsc{multiwoz} & \textsc{sgd} \\
     \midrule
    \multicolumn{3}{c}{Dialogue State Tracking}                   \\
    \midrule
    \textsc{simpletod (baseline) } &  48.7$_{[0.54]}$ / 55.5$_{[0.37]}$  & 31.6$_{[0.45]}$     /  30.2$_{[0.2]}$   \\
    \textsc{generative}  &  31.4$_{[3.17]}$  / 47.9$_{[0.5]}$ & 26.6$_{[0.2]}$    /  31.2$_{[0.38]}$      \\

    \midrule
    \multicolumn{3}{c}{Dialogue Act Prediction}     \\
    \midrule
     \textsc{simpletod (baseline)}    &  76.2$_{[0.31]}$ / 83.69$_{[0.16]}$ & 50.75$_{[0.34]}$  / 50.8$_{[0.13]}$     \\
    \textsc{generative} &   76.2$_{[0.31]}$  / 80.38$_{[0.4]}$ &  47.6$_{[0.06]}$ / 50.76$_{[0.11]}$          \\
    \textsc{autodial}   & 68.6$_{[1.54]}$ /  71.85$_{[0.66]}$   & 45.5$_{[0.46]}$ / 42.1$_{[0.84]}$     \\


    \midrule
    \multicolumn{3}{c}{Dialogue Intent Prediction}         \\
    \midrule
     \textsc{simpletod (baseline)} &      80.2$_{[0.29]}$ / 86.59$_{[0.42]}$           &  69.5 $_{[0.19]}$ /  66.5$_{[4]}$       \\
    \textsc{generative}       & 80.2$_{[0.29]}$ / 84.7$_{[0.64]}$     & 69.5$_{[0.19]}$   /  66.5$_{[4.06]}$    \\

    \textsc{autodial}     &  \textbf{83.2}$_{[0.06]}$  / 82.2$_{[0.07]}$  &  69$_{[0.73]}$ /  \textbf{74.3}$_{[0.21]}$       \\

    \midrule
    \multicolumn{3}{c}{Dialogue Domain Prediction}               \\
    \midrule
    \textsc{simpletod (baseline)}  &   96.2$_{[1.14]}$ / 96.5$_{[1.31]}$  &   75.16$_{[0.66]}$   / 74.5$_{[0.9]}$      \\
    \textsc{generative}        & 94.3$_{[1.02]}$ / 97.16$_{[0.66]}$  &  74.8$_{[0.27]}$ / 73.86$_{[1.97]}$  \\
    \textsc{autodial}  &   \textsc{96.4}$_{[0.15]}$ / \textbf{97.13}$_{[0.06]}$ &  \textbf{78.4}$_{[0.46]}$  / \textbf{78.5}$_{[0.79]}$ \\


    \bottomrule
  \end{tabular}
   \caption{Results demonstrating the impact of Pre-finetuning.
    Numbers in subscript denote standard deviation observed across 3 runs. Bolded numbers represent cases where pre-finetuing helps \textsc{autodial} outperform the two baselines \textsc{simpletod} and \textsc{generative}. Although pre-finetuning seemed to help on intent and domain predictions tasks, the gains did seem substantial enough compared to pre-training.}
  \label{tab:role_of_pft}
\end{table*}

\begin{table}
\vspace{-1em}
\footnotesize
\centering
  \begin{tabular}{lll}
    \toprule

    Decoder     &  Small / Large & Small / Large  \\
                & \textsc{simpletod}  & \textsc{autodial} \\
                & \textsc{(baseline)} & \\
    \toprule
    \multicolumn{3}{c}{\textsc{multiwoz}}                   \\
    \cmidrule(r){1-3}
    DST & 397.12 / 624.87 & -       \\
    Act             & 50.96 / 190.26  &  \textbf{26.8} / 32.31     \\
    Intent        &  83.16 / 127.4 &   \textbf{26.28} / 32.47         \\
    Domain                 &  75.42 / 115.81 &  \textbf{26.92} / 37.71   \\
    \midrule
    \multicolumn{3}{c}{\textsc{google sgd}}                   \\
    \cmidrule(r){1-3}
    DST &  1621.05 / 2286.81       &  - \\
    Act              &   1109.95 / 1800  &    \textbf{196.226} / 243   \\
    Intent         &    408.69 / 545 &   \textbf{198.47}  / 243.19      \\
    Domain                  &    377.8  /  561.12  &   \textbf{195.59} / 239.69          \\
    \bottomrule
  \end{tabular}
  \caption{Inference Time (in seconds) of all decoders of \textsc{autodial}. Inference times for each decoder were averaged out over 10 evaluation runs. Classification decoders are substantially faster than generative decoders used in \textsc{simpletod}.}
    \label{tab:inference_results}
    \vspace{-2em}
\end{table}


\paragraph{Models: }  We trained \textsc{autodial} with two training strategies, fine-tuning and pre-finetuning, the results of which have been shown in \autoref{tab:quant_results} and \autoref{tab:role_of_pft} respectively. We trained with two model sizes for each training stage. The smaller variant has 6 layers each both in the encoder and decoder, thus having $\sim$ 2x fewer parameters compared to the larger one (\autoref{tab:param_count_decoder}). The setup and hyperparameter details have been provided in Appendix.

\paragraph{Pre-finetuning} We pre-finetuned \textsc{bart} on three large dialogue datasets: all three variants of \textsc{taskmaster} \cite{48484}, \textsc{msr e2e} \cite{li2018microsoft}, and \textsc{multidogo} \cite{peskov-etal-2019-multi} for 2.2M training steps. We provide more details of the data in \autoref{sec:pft_data}.

\section{Results}

\paragraph{Performance of \textsc{autodial}}
From Table \ref{tab:quant_results}, we can see that \textsc{autodial} performs comparably on most tasks while being 15x parameter deficient than \textsc{simpletod} (26M vs 406M parameters). We observed that \textsc{autodial} lags behind \textsc{simpletod} by 17.9\% on the Dialogue Act Prediction task from \textsc{multiwoz}. Interestingly, the same is not true on \textsc{google sgd} where \textsc{autodial} outperforms \textsc{Simpletod}  by 2.3\%, despite the larger number of labels \textsc{google sgd} has (4x more). \textsc{autodial} maintained parity with \textsc{generative} and \textsc{simpletod} baselines on both Dialogue Intent Prediction and Dialog Domain Prediction tasks.

\paragraph{Reduction in Training and Inference Time}
\autoref{tab:inference_results} showed that \textsc{autodial} provided significant speedups in both inference and training time compared to both baselines. Specifically, we observed 2-6x reduced inference times on Dialogue Act, Intent, and Domain Prediction tasks of both \textsc{multiwoz} and \textsc{google sgd}. Additionally, \textsc{autodial} required roughly 2.6x less training time compared to \textsc{simpletod} under identical experimental settings.

\vspace{-0.5em}
\paragraph{Does Pre-finetuning help ?}
The benefits of pre-finetuning were primarily seen in the larger variant of \textsc{generative} baseline and \textsc{autodial} (\autoref{tab:role_of_pft}). With the large variant of \textsc{generative} baseline, we saw an improvement of 13.4\% and 11.9\% on Dialog Act Prediction task of \textsc{multiwoz} and \textsc{google sgd} respectively. However the gains were marginal on the Dialog Intent and Domain Prediction task. We had a similar finding with \textsc{autodial} wherein we observed an improvement of 10.5\% and 6.7\% on the Intent and Domain Prediction task of \textsc{google sgd} with pre-finetuning. This observation showed that pre-finetuning provides a moderately better initialization point than pre-training when the encoder weights are kept frozen and that the positive effects of it are likely to become pronounced in larger models.



\paragraph{Effectiveness of the frozen encoder}
As pointed out in \autoref{sec:experimental_details}, the encoder is kept non-trainable in both \textsc{generative} baseline and \textsc{autodial}, we looked at the performance gap across two scenarios, one in which the encoder is trainable and vice versa by looking at performance of \textsc{simpletod} and \textsc{generative} baselines. We observed nearly equivalent performance from both baselines on DST task of \textsc{google sgd}, Dialogue Intent and Domain Prediction tasks of \textsc{multiwoz} and \textsc{google sgd} (shown in \autoref{tab:quant_results}). This indicated that pre-trained knowledge of the encoder can be effectively extended through trainable parallel decoders.




\section{Conclusion}
We presented \textsc{Autodial}, an extension of the \textsc{bart} architecture for task-oriented dialogue systems. \textsc{Autodial} achieved comparable performance to a fully generative approaches such as \textsc{simpletod} while requiring fewer parameters and providing faster inference times. Although this task aimed at studying how classification tasks can benefit from \textsc{autodial}, future work would include how the gains can be realized for generative tasks. Additionally \textsc{autodial} assumes that all the decoders require the same context input. While many dialogue tasks can be completed with user and system responses which is what we use, there are tasks such as end-to-end dialogue generation that require multiple inference steps. To accommodate for such use cases, \textsc{autodial} will require some task-specific modifications.

\paragraph{Ethical considerations / Impact Statement}
As more research is being done with scaling LMs, we believe that it is equally important to study how these LMs can be made more accessible and enabled to be used in real-world scenarios. With \textsc{autodial} we hope to highlight the efficacy of using parallel decoders to perform inference for numerous dialogue tasks while staying competitive and being parameter and memory efficient. We hope to see more work being done in this space especially for restricted compute use-case scenarios. Just like any other LM, special care needs to be given in the collection of data for training and capturing unintended behaviors such as biases or heuristics during deployment.

\bibliography{anthology,custom}
\bibliographystyle{acl_natbib}

\appendix


\section{Setup}
\label{sec:setup}
\textsc{autodial} was implemented using ParlAI \cite{miller2017parlai} and Pytorch \cite{NEURIPS2019_bdbca288}. We conducted all experiments on an NVIDIA A100 GPU, using a batch size of 8. The hyperparameters used in our experiments have been listed in Table \ref{tab:hyperparams}. All of our experiments have been averaged out over 3 runs to account for seed variance.

\section{Data}
\label{sec:pft_data}

\paragraph{TaskMaster: }
The TaskMaster dataset is a large-scale collection of goal-oriented conversations that focuses on simulating human-like dialogues for task-oriented systems. The dataset hopes to provoke interest in written vs spoken language. The datasets consists of two-person dialogs, The dataset was created using the Wizard-of-Oz methodology, where human participants were given specific roles to play while generating the conversations.

\paragraph{MSR-E2E: } MSR-E2E is a dataset of human-human conversations in which one human plays the role of an Agent and the other one plays the role of a User. It contains human-annotated conversational data in three domains (movie-ticket booking, restaurant reservation, and taxi booking).

\paragraph{Multidogo}: MultiDoGo is a large task-oriented dataset collected in a Wizard of Oz fashion, using both crowd and expert annotators with annotations at varying levels of granularity. It contains 81K dialogues harvested across six domains. It was crafted using the Wizard-of-Oz approach wherein a crowd-sourced worker (the “customer”) is paired with a trained annotator (the “agent”).


\begin{table}
\small
  \begin{tabular}{lll}
    \toprule
    \multicolumn{2}{c}{Generation}                   \\
    \cmidrule(r){1-3}
    Component     &  Hyperparameter  \\
    \midrule
     Learning Rate (\textsc{generator} \textsc{multiwoz})  & 2e-5  &       \\
     Learning Rate (\textsc{generator} \textsc{sgd}) & 1e-5 \\
     Learning Rate (\textsc{simpletod} \textsc{multiwoz})  & 2e-5  &       \\
     Learning Rate (\textsc{simpletod} \textsc{sgd}) & 1e-5 \\
     Learning Rate (\textsc{autodial})  & 7e-5  &       \\
     Pre-finetuning LR & 1e-6 \\
     Sequence Length & 1000 \\
     Optimizer & AdamW \\
     Batchsize & 8 \\
     Eval batch size & 16 \\
     Warmup updates & 100 \\
     Dropout & 0 \\
     Gradient clipping & 1.0 \\
     Num epochs & 5 \\

    \bottomrule
  \end{tabular}

   \caption{Hyperparameters used for our experiments}
    \label{tab:hyperparams}
\end{table}


\end{document}